\documentclass[conference]{IEEEtran}
\IEEEoverridecommandlockouts
\usepackage{cite}
\usepackage{amsmath,amssymb,amsfonts}
\usepackage{algorithmic}
\usepackage{subcaption}
\usepackage{graphicx}
\usepackage{textcomp}
\usepackage{xcolor}
\usepackage{cleveref}
\usepackage{todonotes}

\usepackage[detect-none]{siunitx}
\newcommand{\mAh}{mAh}
\newcommand{\mps}{\meter / \second}
\usepackage{glossaries}
\newacronym{tof}{ToF}{time-of-flight}
\newacronym{ekf}{EKF}{extended Kalman filter}
\newacronym{fov}{FoV}{field-of-view}
\newacronym{soc}{SoC}{system-on-chip}
\newacronym[plural=MCUs,firstplural=microcontroller units (MCUs)]{mcu}{MCU}{microcontroller unit}
\newacronym[plural=CNNs,firstplural=convolutional neural networks (CNNs)]{cnn}{CNN}{convolutional neural network}
\newacronym{slam}{SLAM}{simultaneous localization and mapping}
\newacronym{lidar}{LiDAR}{light detection and ranging}
\newacronym{radar}{RADAR}{Radio Detection and Ranging}
\newacronym{iqd}{IQD}{interquartile deviation}
\newacronym{bce}{BCE}{binary cross-entropy}
\newacronym{auroc}{AUROC}{area under receiver operating characteristic curve}
\newacronym{rmse}{RMSE}{root mean squared error}
\newacronym{uart}{UART}{universal asynchronous receiver/transmitter}
\newacronym{imav}{IMAV}{International Micro Air Vehicles}
\newacronym{soa}{SoA}{state of the art}
\newacronym{uav}{UAV}{unmanned aerial vehicle}
\newacronym{vml}{VML}{visual model‐predictive localization}
\newacronym{dnn-rl}{DNN-RL}{deep neural network reinforcement learning}
\newacronym{ssd}{SSD}{single-shot multibox detector}
\newacronym{vio}{VIO}{visual-inertial odometry}
\newacronym{qat}{QAT}{quantization aware training}
\newacronym{oa}{OA}{obstacle avoidance}
\newacronym{cf}{CF}{Crazyflie 2.1}
\newacronym{us}{US}{ultrasound}
\newacronym{aln}{AlN}{Aluminum Nitride} 
\newacronym{pmut}{PMUT}{piezoelectric micromachined ultrasonic transducer} 
\newacronym{mems}{MEMS}{micro-electromechanical systems}
\newacronym{pulp}{PULP}{Parallel Ultra-Low-Power}
\newacronym{odr}{ODR}{output data rate}
\newacronym{iq}{IQ}{in-phase and quadrature}
\newacronym{snr}{SNR}{signal-to-noise ratio}

\def\BibTeX{{\rm B\kern-.05em{\sc i\kern-.025em b}\kern-.08em
    T\kern-.1667em\lower.7ex\hbox{E}\kern-.125emX}}
\begin{document}

\title{BatDeck: Advancing Nano-drone Navigation with Low-power Ultrasound-based Obstacle Avoidance\\

\thanks{Identify applicable funding agency here. If none, delete this.}
}
\author{
\IEEEauthorblockN{Hanna M\"uller\IEEEauthorrefmark{1}, Victor Kartsch\IEEEauthorrefmark{1}, Michele Magno\IEEEauthorrefmark{1}, Luca Benini\IEEEauthorrefmark{1}\IEEEauthorrefmark{2}}

\IEEEauthorblockA{\IEEEauthorrefmark{1}Integrated Systems Laboratory / Center for Project-Based Learning  - ETH Z\"urich, Switzerland}

\IEEEauthorblockA{\IEEEauthorrefmark{2}DEI - University of Bologna, Italy}

Email: \{hanmuell, vkartsch, mmagno, lbenini\}@ethz.ch
}

\maketitle

\begin{abstract}
Nano-drones, distinguished by their agility, minimal weight, and cost-effectiveness, are particularly well-suited for exploration in confined, cluttered and narrow spaces. Recognizing transparent, highly reflective or absorbing materials, such as glass and metallic surfaces is challenging, as classical sensors such as cameras or laser rangers often do not detect them. 
Inspired by bats, which can fly at high speeds in complete darkness with the help of ultrasound, this paper introduces \textit{BatDeck}, a pioneering sensor-deck employing a lightweight and low-power ultrasonic sensor for nano-drone autonomous navigation. This paper first provides insights about sensor characteristics, highlighting the influence of motor noise on the ultrasound readings, then it introduces the results of extensive experimental tests for obstacle avoidance (OA) in a diverse environment. Results show that \textit{BatDeck} allows exploration for a flight time of 8 minutes while covering 136m on average before crash in a challenging environment with transparent and reflective obstacles, proving the effectiveness of ultrasonic sensors for OA on nano-drones.

\end{abstract}

\begin{IEEEkeywords}
nano-drone, nano-UAV, ultrasound, obstacle avoidance, autonomous navigation
\end{IEEEkeywords}

\section{Introduction}

The global drone market, with a projected value of \$54.6 billion by 2030\cite{statista}, reflects this technology's rapid growth and increasing relevance due to its versatility for various applications, ranging from agriculture, construction, and public safety to urban planning. While larger drones are known for their capability to execute complex tasks, nano-drones offer unique advantages, especially in constrained environments like greenhouses and buildings. Their small size (typically $\sim$\qty{10}{\centi\meter} diameter~\cite{HASSANALIAN201799}) makes them safer for people and property. Given its versatility and modularity, the \gls{cf}, an open software, open hardware nano-drone, has gained popularity among research and commercial entities~\cite{zhang2024endtoend,lamberti2022tiny,mueller2023robust,mcguire2019minimal,10423569}.

A critical aspect of ensuring the drone platform's safe and efficient operation is \gls{oa}. \gls{oa} requires appropriate sensing technologies that typically include \gls{lidar}, \gls{radar}, cameras~\cite{zhang2024endtoend, 10423569, lamberti2022tiny}, and laser-based \gls{tof} sensors \cite{10423569, mueller2023robust}. However, \gls{lidar} and \gls{radar} are bulky and power-hungry, which forbid their use for nano-drones applications, while cameras and laser-based \gls{tof} sensors do not perform well when dealing with reflective surfaces or transparent barriers such as glass walls. Although sensor fusion of multiple powerful onboard sensors allows larger drones to overcome inaccurate measurements, nano-drones, typically below \qty{50}{\gram}, must resort to lightweight and low-power sensors due to their payload and power limitations~\cite{HASSANALIAN201799}. Taking inspiration from the echolocation techniques employed by bats and dolphins, \gls{us} sensors, which measure the \gls{tof} in echos of high-frequency sound, are a promising alternative, offering better performance when dealing with sound-reflecting objects independently of color, transparency, or texture~\cite{laurijssen2019flexible,forouher2016sensor}. 

A wide range of ultrasonic sensors applicable for use in robotics are currently available on the market and have been examined in the literature\cite{zhmud2018application, rshen2019new}. However, their size, weight, and energy requirements preclude their use in nano-drone applications. The recently introduced ICU-30201 sensor from TDK overcomes these limitations with its ultra-low power consumption (below \qty{1}{\milli\watt}) and compact form factor (\qty{5.17}{\times}\qty{2.68}{\times}\qty{0.9}{\milli\meter}), making it an ideal choice for integration into small-scale robotic platforms~\cite{przybyla2023mass}. 
\begin{figure}
\centering
\includegraphics[width=0.95\columnwidth]{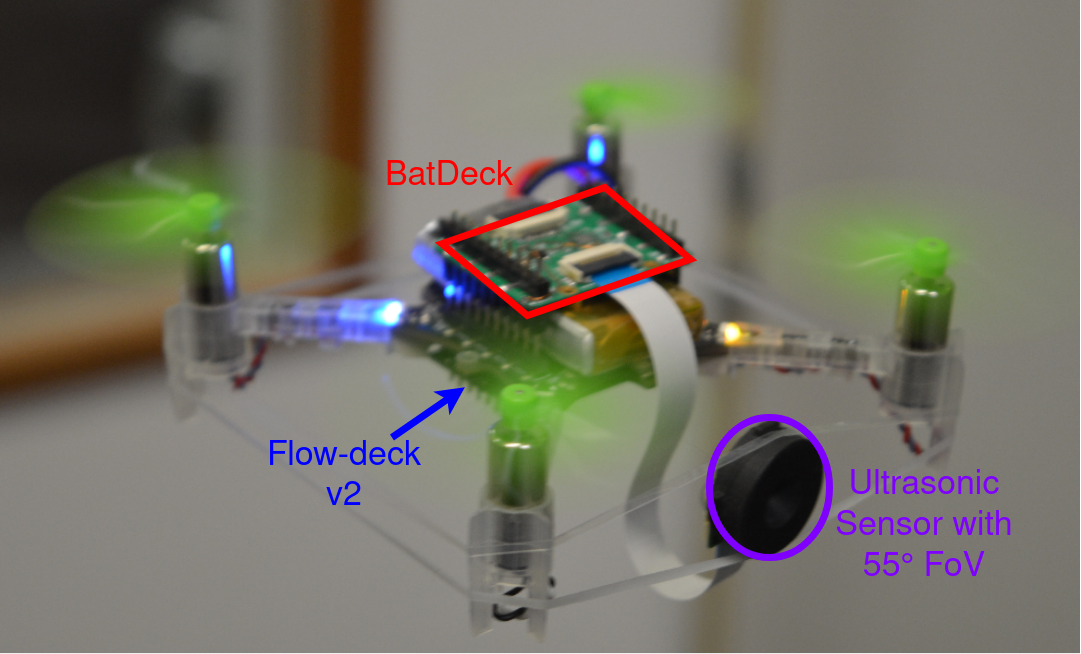}
\caption{The Crazyflie 2.1 with the \textit{BatDeck} and the Flow-deck v2, during an \gls{oa} test flight.}
\label{fig:hw_overview}
\end{figure}
Exploiting this novel \gls{us} sensor, we move nano-drones toward bat-like navigation with our four main contributions:

(1) Design and implementation of an \gls{us} extension for nano-drones:
This contribution focuses on the design and characterization of an \gls{us}-based perception system for nano-drones (shown in \Cref{fig:hw_overview}). We integrate the compact and energy-efficient ICU-30201 sensor into the drone's architecture, enabling it to emit and receive ultrasonic signals for obstacle detection and navigation, akin to the natural echolocation used by bats.

(2) Characterization and countermeasures against motor noise:
Vibrations affect \gls{us} sensors. Therefore, we characterize the noise and evaluate filtering methods to extract the echolocation signal. We also characterize the \gls{us} obstacle detection performance at different distances and angles towards concrete and glass surfaces.

(3) Implementation and evaluation of an \gls{oa} algorithm tailored for resource-constrained hardware (ARM Cortex-M4) onboard a nano-drone in terms of power, latency, and memory requirements.
 
(4) In-field evaluation and comparison with laser-based \gls{tof} sensors in \gls{oa} applications:
The final section presents an in-field evaluation of the nano-drone equipped with the \textit{BatDeck}, comparing its \gls{oa} capabilities against a more conventional laser-based \gls{tof} system. This analysis focuses on effectiveness, power consumption, and adaptability to various environments, providing insights into the advantages and potential limitations of \gls{us} versus laser-based \gls{tof} sensors in real-world applications and \gls{oa} scenarios.

\section{Related Work}
\gls{oa} can rely on several sensing sources, including \gls{lidar}, \gls{radar}, cameras, laser-based \gls{tof}, and \gls{us} sensors. \gls{lidar} sensors reconstruct a 3D space using consecutive range measurements projected in all directions sequentially. While \gls{lidar} technology is one of the best solutions for \gls{oa}, its power consumption and size are incompatible with nano-drones.

Cameras have also been used successfully for \gls{oa} tasks. Tiny, low-power cameras can easily be mounted and operated on small systems such as nano-drones. For instance, authors in \cite{palossi201964} designed a vision-based \gls{cf}-compatible shield for \gls{oa}. The device integrates a low-power monochrome QVGA camera (Himax HM01B0) and GAP8, an \gls{pulp} \gls{soc} for efficient computation. The navigational commands are based on PULP-DroNet, a convolutional neural network for \gls{oa} fine-tuned for this platform. Authors in ~\cite{zhang2024endtoend} use a lightweight CNN depth estimation network for \gls{oa} - the generated depth map could even be used for path planning. While camera-based approaches are compatible with nano-drones, they still lack performance in poor lighting conditions or when navigating across an environment with objects that have not been included in the training sets.

Authors in \cite{10423569} have recently presented Stargate, a multimodal sensor fusion system for nano-drones, enabling autonomous navigation and gate detection through sensor fusion between a low-power grayscale camera (Himax HM01B0) and a multi-pixel (8x8) laser-based \gls{tof} (VL53L5CX) sensor from STMicroelectronics. The system is trained entirely on synthetic data and tested on a \gls{cf} drone, demonstrating robust navigation with a low failure rate in various scenarios. While this approach is working in many environments, its robustness can be improved with fusing an additional sensor perceiving glass or visually reflective materials correctly.

Attempting to take advantage of sound, which can also detect visually transparent and shiny obstacles, Duembgen et al. \cite{duembgen2023blind} developed a real-time, model-based system for auditory localization and mapping tailored for compact robots. Using buzzers and basic microphones, the system achieves centimeter precision without prior calibration or training. However, applying this method onboard nano-drones presents challenges due to the constraints related to weight and size, as well as interference with audible motor noise, rendering it impractical for such applications.

Recent advancements in \gls{mems} design have enabled the development of new technologies that are low-power, lightweight, and compact. The ICU-30201 ultrasonic \gls{tof} incorporates \gls{aln} \glspl{pmut} to offer enhanced sensitivity and reduced power consumption when compared to traditional piezoceramic transducers. Furthermore, its size and weight are compatible with nano-drones.

In this paper, we leverage the capabilities of the TDK ICU-30201 to develop \textit{BatDeck}, an expansion board compatible with \gls{cf} drones, enhancing their performance for \gls{oa} tasks. We detail the sensor's characterization under various flight conditions and across different materials and provide a quantitative assessment of its efficacy in complex environments compared to conventional laser-based \gls{tof} sensors.

\section{Hardware Setup}
In this section, we introduce the ICU-30201, the \gls{us} sensor used in this paper, as well as details on the \textit{BatDeck}, our custom-designed expansion deck. In the following, we also introduce details about our flying platform, the \gls{cf} drone.

\subsection{ICU-30201}
The ICU-30201 is a miniature, ultra-low power \gls{mems}-based ultrasonic \gls{tof} transceiver. It integrates a nominally $f_{op} =50kHz$ \gls{pmut} and a \qty{40}{\mega\hertz} CPU for sampling and pre-processing.
It can record maximally 340 \gls{iq} data samples per measurement, converted to baseband. The \gls{odr} can be configured as $\gls{odr} = f_{op}/N$ where $N = 2,4,8$. The \gls{iq} allows computing phase and magnitude of the reflected \gls{us} waves.
\subsection{BatDeck}
This section presents the design and implementation of the \textit{BatDeck}, a custom-made extension deck for the Crazyflie, which features four slots for attaching TDK ICU-x0201 sensors. In this work, we only use one slot to attach a forward-facing ICU-30201 ultrasonic sensor with a \qty{55}{\degree} \gls{fov} horn, attached to the drone with two rubber bands (for dampening). The \textit{BatDeck} with one sensor weighs in total \qty{3}{\gram}, where the expansion deck PCB itself contributes 1.37 g, the sensor horn \qty{1.25}{\gram}.
\subsection{Crazyflie Platform}
This work employs the \gls{cf} from Bitcraze, a 10-centimeter nano-drone featuring an STM32F405 \gls{mcu}, which oversees state estimation and actuation control. The drone is also equipped with a Flow-deck v2, which enables optical flow and height measurements to improve state estimation. We use 7$\times$\qty{16}{\milli\meter} 19000 KV motors from betafpv, 47-17 propeller from Bitcraze, and an \qty{350}{\mAh} LiPo battery from Tattu. This base configuration weighs \qty{34}{\gram}.

\section{Sensor and Noise Characterization and Calibration}
\label{lab:charact}
This section presents a performance evaluation of the ultrasonic sensor in the presence of motor noise (during flight), maximum range, and measurement performance for angled obstacles.
Transmission/Reception commands can be configured in cycles of the MUTCLK, which operates at $16\times f_{op}$. We configure the sensor to transmit for 512 cycles and actively dampen ringdown artifacts for an additional 45 cycles. We chose $N = 4$, giving us a range of \qty{4.6}{\meter} and a resolution of \qty{1.35}{\centi\meter} (with $f_{op} = 50kHz$ and 340 samples per measurement). Note that $f_{op}$ can change from sensor to sensor and in different environmental conditions.

Measurement acquisition runs at \qty{33}{\hertz}, however, the radio-based transmission is limiting the logging frequency. Considering that the phase information is not used in this paper, we directly stream out magnitude data (instead of transmitting \gls{iq} parameters), at a measurement rate of $\sim$\qty{4.8}{\hertz}. The magnitude data corresponds to the intensity of the received \gls{us} wave, which we can use to detect sound-reflecting objects and calculate the distances to them. As the drone's motor vibrations lead to noise on the sensor data, we first characterize the noise induced by the motors and later evaluate the sensor performance.

\subsection{Motor Noise}
\label{lab:motor_noise}
To characterize the motor noise, we configure the ICU-30201 to only receive (i.e., no \gls{us} waves are emitted) while the drone hovers at \qty{0.7}{\meter}. 

The goal is to increase \gls{snr} by filtering out the motor noise to be able to detect the reflected signal from the closest obstacle, while choosing an algorithm with low computational and memory requirements for onboard execution. 
Filtering can be applied on both slow-time (i.e., over consecutive \gls{us} measurements) or fast-time (i.e., over the consecutive 340 \gls{us} samples from the same measurement).

To save memory, we use an exponentially moving average in slow-time: $y_{i_{n}} = \frac{K_s - 1}{K_s}y_{i_{n-1}} + \frac{1}{K_s}x_{i_{n}}$ where $n$ denotes the measurement number and $i$ the sample number within the measurement, $y$ denotes the filtered magnitude while $x$ is the raw measurement. In fast-time, we use a averaging filter: $y_{i_{n}} = \sum_{j=i-K_f+1}^{i}\frac{x_{j_{n}}}{K_f}$. 

\begin{figure}
\centering
\includegraphics[width=\columnwidth]{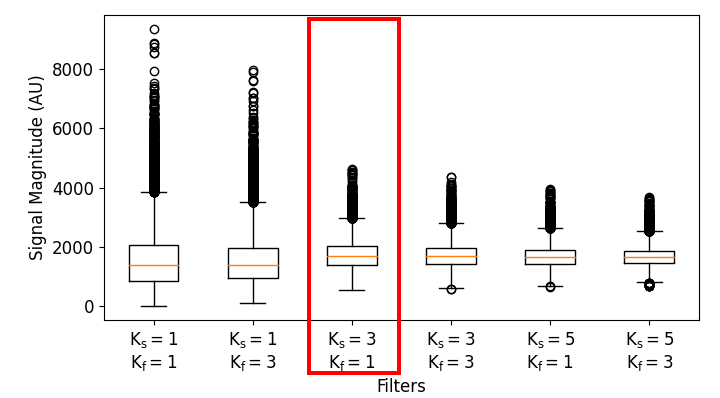}
\caption{The effect of different filter lengths in slow and fast time to the motor noise standard deviation.}
\label{fig:motor_var}
\end{figure}

In \Cref{fig:motor_var}, we show the standard deviation, median, and outliers of the motor induced noise on the sensor signal. We evaluate the unfiltered baseline ($K_s=1, K_f=1$) against $K_f=3$, $K_s=3$ and $K_s=5$. As shown in \Cref{fig:motor_var}, filtering along the slow-time reduces the noise variance significantly. We pay in a time delay for filtering along slow-time, while for filtering along fast-time, we compromise the resolution of the measured distance to the detected objects. 

We chose an exponential moving average with $K_s=3$ and decided against fast time filtering ($K_f=1$), as the additional improvements are minor compared to the costs. Compared to the baseline, the standard deviation of the motor noise is reduced by \qty{50}{\percent}, and the mean remains as unfiltered at \qty{\sim1700}{AU}.

\begin{figure}
\centering
\includegraphics[width=0.6\columnwidth]{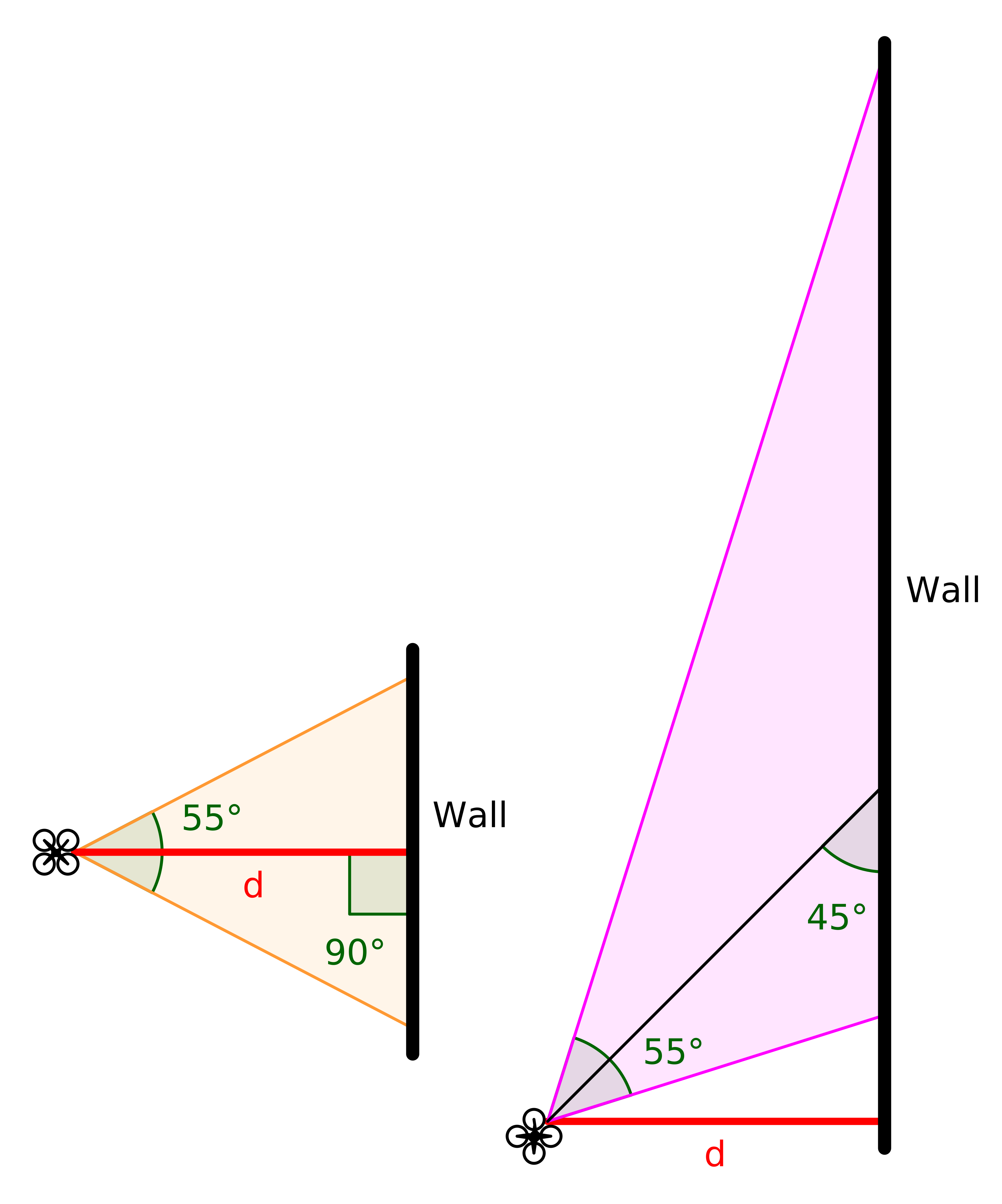}
\caption{The test setup, on the left, the drone is facing the wall at \qty{90}{\degree}, on the right at \qty{45}{\degree}. The distance to the wall ($d$) is marked in red in both scenarios.}
\label{fig:charac_setup}
\end{figure}

\subsection{Sensor Characterization}
\Cref{fig:charac_setup} shows our experimental setup for the sensor characterization. We command the drone to fly at the target height of \qty{0.7}{\meter}, pointing towards a wall. The sensor is positioned pointing perpendicular (\qty{90}{\degree}) or in \qty{45}{\degree} towards the wall. The wall is wide enough to span the whole sensor \gls{fov} (\qty{55}{\degree}). However, multipath reflections over the ground are possible and can lead to additional signal peaks.

\begin{figure}
\centering
\includegraphics[width=0.9\columnwidth]{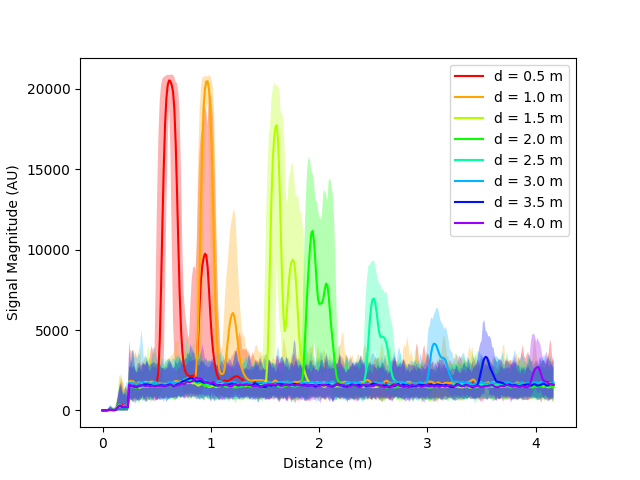}
\caption{Mean and min/max (shaded) for \qty{0.5}{\meter} to \qty{4}{\meter} distance to the wall at \qty{90}{\degree}.}
\label{fig:charac_90deg}
\end{figure}

During our experiments, we sweep the distance $d$ (shown in \Cref{fig:charac_setup}) between the sensor and the wall in the range \qty{0.5}{\meter} – \qty{4}{\meter} with a step of \qty{0.5}{\meter}. For each step, we acquire 100 measurements. We use the onboard exponential moving average with $K_s=3$, shown in \Cref{lab:motor_noise}. Note that all measurements are acquired on a flying drone, leading to slight inaccuracies in position/attitude control over time.
\Cref{fig:charac_90deg} shows the average and min/max (shaded) of the \qty{90}{\degree} measurements. As expected, we see the noise floor at around 1700 AU and clear signal peaks of the wall for each distance measured.  

\begin{figure}
\centering
\includegraphics[width=0.9\columnwidth]{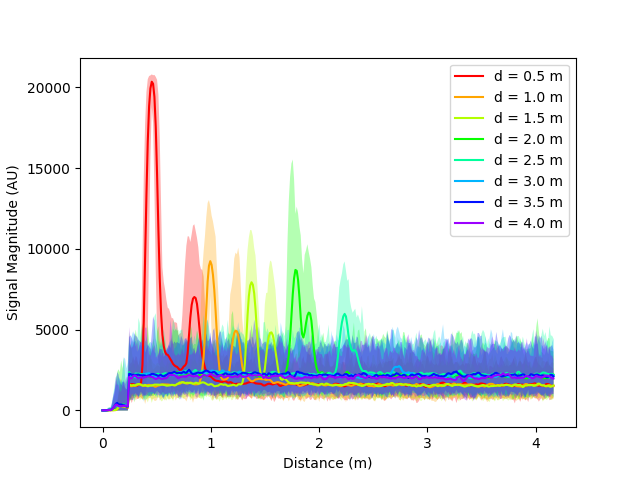}
\caption{Mean and min/max (shaded) for \qty{0.5}{\meter} to \qty{4}{\meter} distance to the wall at \qty{45}{\degree}.}
\label{fig:charac_45deg}
\end{figure}

In the \qty{45}{\degree} scenario in \Cref{fig:charac_45deg}, we see that the reflection causing the main peak is coming from the closest point to the sensor on the wall and not from the center of the sensor \gls{fov}. This is due to destructive interference from multiple echos along the wall superimposed on the receiving sensor.

\begin{figure}
\centering
\includegraphics[width=0.9\columnwidth]{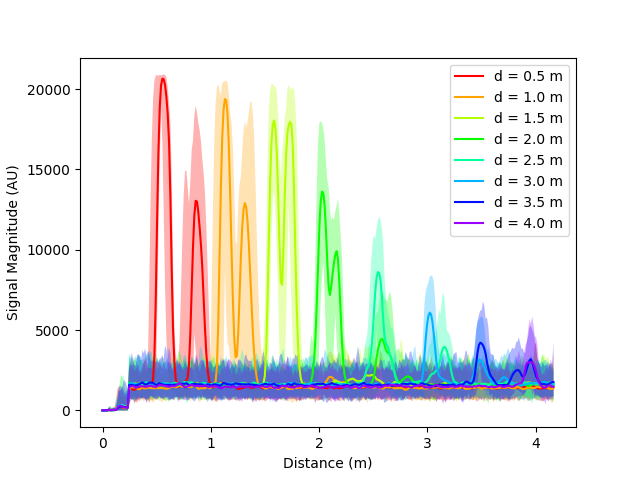}
\caption{Mean and min/max (shaded) for \qty{0.55}{\meter} to \qty{4}{\meter} distance to a glass front at \qty{90}{\degree}.}
\label{fig:charac_90deg_glass}
\end{figure}

We repeated the \qty{90}{\degree} experiment with a glass front to assess the sensor performance towards other materials. In \Cref{fig:charac_90deg_glass} the \gls{us} sensor performs similarly on glass as on concrete. 

\section{\gls{oa} Algorithm Implementation}
\begin{figure}
\centering
\includegraphics[width=0.45\columnwidth]{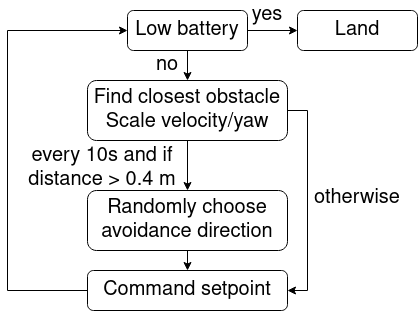}
\caption{Algorithmic flow of the \gls{oa} implementation.}
\label{fig:flow}
\end{figure}
To evaluate the performance of the \gls{us} sensor in an \gls{oa} use case, we have designed and implemented an \gls{us}-\gls{oa} algorithm that can run directly on the ARM Cortex-M4 core on the \gls{cf}, as described in \Cref{fig:flow}. The drone scales the velocity and yaw rate according to the distance to the closest obstacle. For random exploration, the yaw rate sign (avoiding obstacles to the left or right) is randomly chosen every \qty{10}{\second} - besides, when an obstacle is detected closer than \qty{40}{\centi\meter}, the direction is not changed to avoid turning back once the obstacle is almost avoided and staying longer than necessary close to obstacles.

The ultrasonic wave loses intensity over time, calling for lowering the threshold for object detection for longer distances. However, as every reflection but the one of the closest obstacle is irrelevant to our \gls{oa} algorithm (double reflections are never the closest intensity peaks and therefore do not have to be filtered out) and we do not experience higher noise at closer distances, we decide to use a constant threshold. 
As shown in \ref{lab:charact}, all the (measured) noise is contained in values up to $\sim$6000, so we choose $Thres_{const} = 6000$.

\begin{figure}
\centering
\includegraphics[width=0.9\columnwidth,trim={0.42cm 0.7cm 0.92cm 0.9cm},clip]{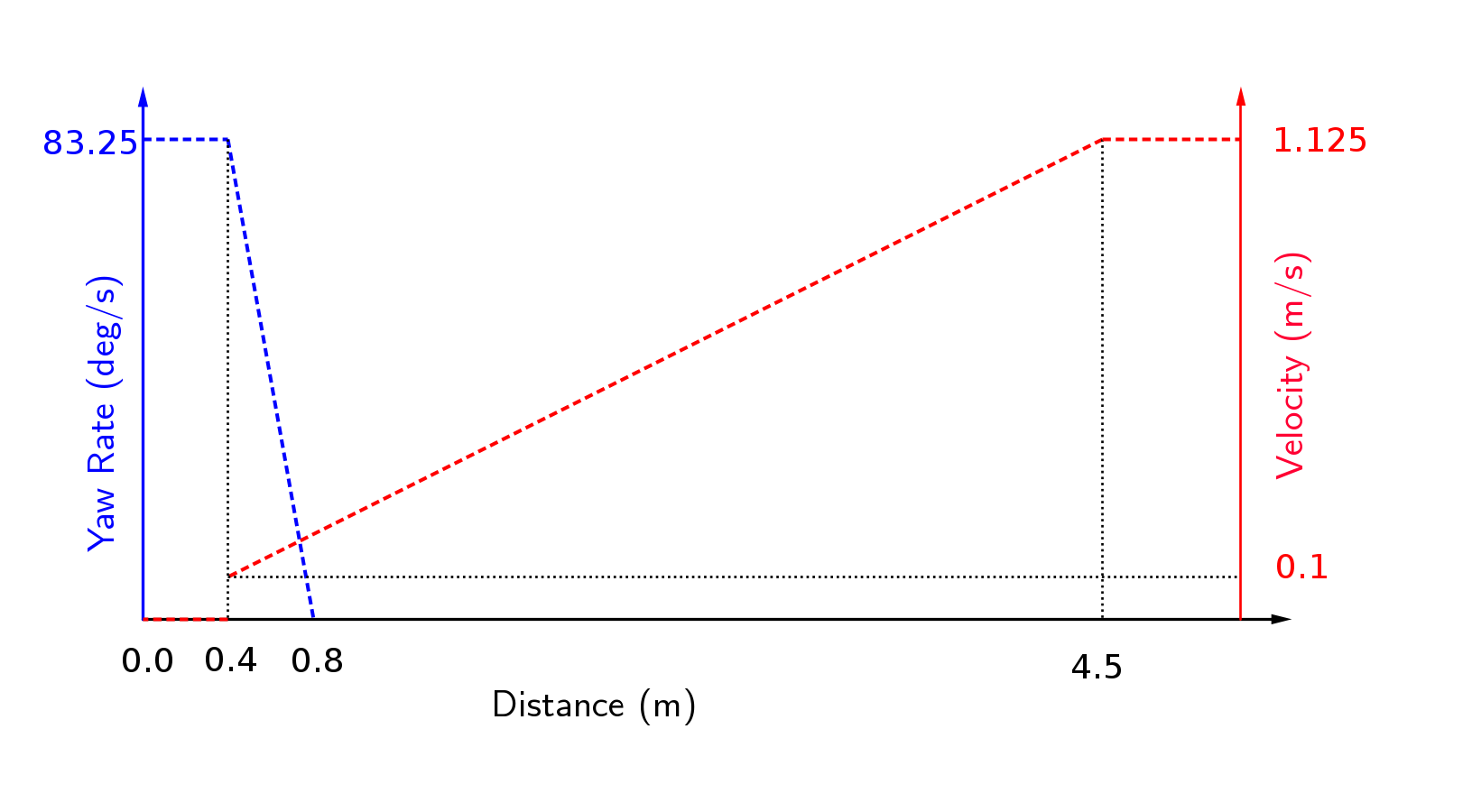}
\caption{Velocity and yaw rate scaling according to distance to the closest object.}
\label{fig:vel_scaling}
\end{figure}

Once we determine the distance to the closest object, we scale the velocity in the x direction and the yaw rate accordingly. For objects detected closer than \qty{40}{\centi\meter}, we stop and only rotate with a yaw rate of \qty{83.25}{\degree\per\second} (leading to a change of \qty{2.5}{\degree} in between measurements). We scale the yaw rate linearly for objects detected between \qty{40}{\centi\meter} and \qty{80}{\centi\meter} away, not turning anymore from \qty{80}{\centi\meter} on. For distances further than \qty{40}{\centi\meter} away, we set the velocity to the distance divided by 4 per second, resulting in a maximum velocity of \qty{1.15}{\mps}. The scaling is visualized in~\Cref{fig:vel_scaling}. To avoid sudden accelerations, we limit the velocity increase to maximally \qty{0.05}{\mps} per measurement (equalling 1 $m/s^2$). Velocity decrements are unrestricted. 
\section{Experimental Evaluation}
In this section, we first evaluate the power consumption, latency, memory requirements, and computational load of the ICU-30201 and the proposed \gls{oa} algorithm running onboard the \gls{cf} on the Cortex-M4 core. Following, we describe the in-field \gls{oa} evaluation with the ICU-30201 sensor and finally compare the results to the performance of a low-power laser-based \gls{tof} sensor (VL53L1) employed for the same task, also investigating causes for failure in both sensors. 
\subsection{Power, Latency and Computational Load}
\begin{figure}
\centering
\includegraphics[width=1\columnwidth]{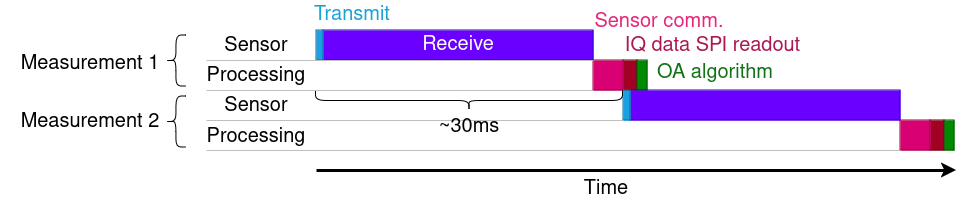}
\caption{Latency of transmission, reception, sensor communication, \gls{iq} data SPI readout and algorithm execution.}
\label{fig:timing}
\end{figure}
For traveling twice the maximum range of 340 samples at $\gls{odr} = f_{op}/4$ in air (equalling $\sim$\qty{4.6}{\meter}), the ultrasonic wave needs $\sim$\qty{27}{\milli\second}. As we currently transmit the whole \gls{iq} data, \qty{1.4}{\milli\second} are needed for the SPI data transfer (\qty{12}{\mega\hertz} SPI is used). An additional $\sim$\qty{3}{\milli\second} are used for the sensor communication before the readout (including waiting for the task to be executed by the FreeRTOS scheduler, adding a variance of +/-$\sim$\qty{1}{\milli\second}) and $<$ \qty{1}{\milli\second} for executing the \gls{oa} algorithm. However, during the SPI transfer and \gls{oa} algorithm, the next measurement can already start as visualized in~\Cref{fig:timing}, leading to a total latency of $\sim$\qty{30}{\milli\second} from the start of a measurement and hence a frequency of \qty{33}{\hertz}. As transmit and receive time are physically limited and can only be influenced by reducing the signal duration and range, only the \qty{3}{\milli\second} for the initial sensor communication could be reduced to influence the maximal measurement rate.

As we use an exponential moving average, the \gls{oa} algorithm can directly operate on the original \gls{iq} data array, and only the 20 ringdown sample magnitudes are saved until the next measurement, leading to a memory requirement of $340\times2\times2 + 20\times2 =$ \qty{1.4}{kBytes} for the samples in int16 format.

The \gls{oa} task adds \qty{2.5}{\percent} to the STM32F405's computational load, still leaving $>$ \qty{57}{\percent} idle. One ICU-30201 sensor consumes $<$\qty{1}{\milli\watt}, which is negligible on a drone using $\sim$\qty{11}{\watt}. 
\subsection{In-field \gls{oa} Tests}
We conducted 10 test flights in an office environment to verify our setup. The office features several desks, chairs, cupboards, glass doors, and other obstacles (a cloth rack, a bicycle, etc.). We show the results in \Cref{tab:oa_us}; the drone successfully flew until the battery ran low in \qty{50}{\percent} of the tests, while it crashed in the others. The causes for the crashes were always either a soft chair or flying just below a table. The average covered distance is \qty{68}{\meter}, and the average flight time is 4'22". The average time/distance until a crash (excluding battery changes) is 8'43"/\qty{136}{\meter}. 
\begin{table}[t]
\caption{Results of 10 random exploration \gls{oa} flights in an office environment.}
\label{tab:oa_us}
\centering
\footnotesize
\begin{tabular}{c|c|c|c|c}
\hline\hline
{\textbf{Exp.}} & \textbf{Time [s]} & {\textbf{Crash}} & \textbf{Distance [m]} \\ 
\hline\hline
(1) & 430 & $\times$    & 107  \\ \hline
(2) & 294 & \checkmark  & 72  \\ \hline
(3) & 392 & $\times$    & 120  \\ \hline
(4) & 394 & $\times$    & 115  \\ \hline
(5) & 188 & \checkmark  & 55  \\ \hline\hline
(6) & 287 & $\times$    & 63  \\ \hline
(7) & 35 & \checkmark   & 8.5  \\ \hline
(8) & 97 & \checkmark   & 27  \\ \hline
(9) & 383 & $\times$    & 90  \\ \hline
(10) & 115 & \checkmark & 27  \\ 
\hline\hline
\textbf{Average} & \textbf{262} & \textbf{50\%} & \textbf{68} \\ \hline\hline
\end{tabular}
\end{table}

\subsection{Comparison to VL53L1}
We repeated the real-world \gls{oa} test with a VL53L1 sensor instead of the ICU-30201, operating at the same frequency (\qty{33}{\hertz}) and using the same avoidance algorithm. Out of 10 flights, \qty{100}{\percent} were ended by a crash, traveling an average distance of \qty{4}{\meter} and time of \qty{9}{\second}. It is to note that the VL53L1 has not only the disadvantage of trouble with highly reflective and absorptive obstacles such as glass/black surfaces but also a more narrow \gls{fov} (\qty{27}{\degree} versus \qty{55}{\degree}), which makes it almost impossible for it to perceive all obstacles it could collide with. However, most crashes were caused by glass doors.

\begin{figure}
\centering
\includegraphics[width=1\columnwidth,trim={0cm 14cm 0cm 0cm},clip]{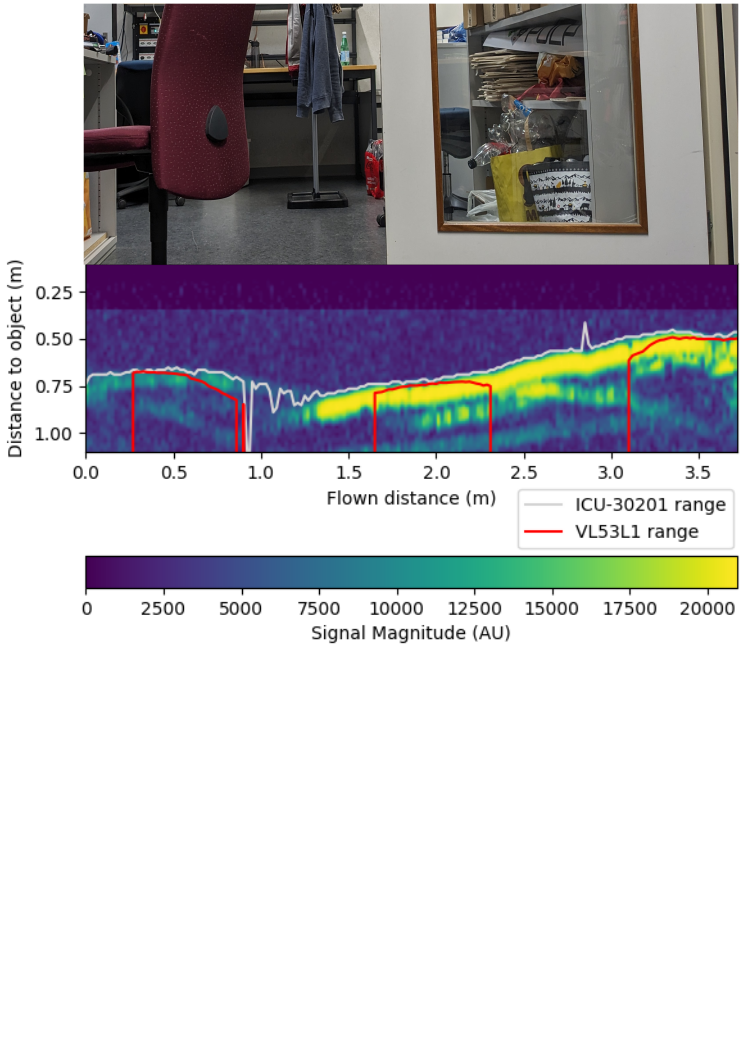}
\caption{ICU-30201 and VL53L1 comparison, both concurrently logged while flying at \qty{0.1}{\mps} from the left to the right while facing the obstacles shown in the top part.}
\label{fig:comp}
\vspace{-2mm}
\end{figure}

To showcase the different perceptions of the two sensors and investigate common causes of failure, we flew along a row of different objects/materials, as displayed in \Cref{fig:comp}. Under the image of the scene, we show the \gls{us} spectrogram and the predicted distances to the closest object from both the ICU-30201 and the VL53L1 sensors. We see the effect of the more narrow \gls{fov} of the VL53L1, enabling it to see more narrow gaps but also provoking a more risky flight behavior. While the VL53L1 fails to see the glass door and the first part of the chair, where it does not cover much of the \gls{fov}, we also note that while the ICU-30201 successfully detects the soft chair, it receives a much weaker reflection from it than from solid obstacles, especially when it is only at the edge of the \gls{fov}.  

In terms of power consumption, the VL53L1 consumes $\sim$\qty{50}{\milli\watt}, while the ICU-30201 consumes $<$\qty{1}{\milli\watt}. However, the \textit{BatDeck} with one sensor weighs \qty{3}{\gram}, while the Multi-ranger deck used for the Vl53L1 experiments reduced to 1 sensor weighs only \qty{2}{\gram}. Both designs can accommodate up to 4 sensors and, therefore, could be optimized for less weight.
\section{Conclusion}
We presented and evaluated the \textit{BatDeck}, featuring a novel, lightweight, and low-power ultrasonic sensor for expanding nano-drones. We provide a motor noise characterization and propose a filter to counteract the motor-induced vibration noise. To further characterize the sensor, we provide measurements at different distances, angles, and of different materials, proving the sensor's capabilities to detect obstacles up to \qty{2.5}{\meter} reliably despite motor noise. 

We show the \gls{oa} capabilities by implementing a proof-of-concept algorithm that can run at \qty{33}{\hertz} and outperforms a classical laser-based \gls{tof} sensor, offering \qty{50}{\percent} higher mission success rate while tracking \qty{17}{\times} more distance. An average flight distance of \qty{136}{\meter} and time of 8'43'' until crash proves the effectiveness of \gls{us} sensors for \gls{oa} on nano-drones.

This work lays the foundation for the use of ultrasonic sensors on nano-drones. Future work aims to implement a self-calibration mechanism for automatic compensation of ringdown effects (improving perception of close obstacles) and motor noise. As the ICU-30201 also features a CPU, the \gls{oa} algorithm could also be moved to the sensor \gls{soc}, lowering the SPI bus load and the latency.

Similarly, we aim to pursue adding multiple ultrasonic sensors, which will allow a full spatial perception of obstacles and, with this, more advanced \gls{oa} algorithms. Also, using phase information and modulating the emitted \gls{us} wave hold potential for improved spatial perception.

Lastly, we aim to explore sensor fusion techniques to integrate information from ultrasonic and laser-ranged \gls{tof} sensors, aiming to increase the robustness of the system when performing exploration with \gls{oa} in mixed scenarios.  

\section*{Acknowledgment}
We thank TDK for their support.
We also thank Christian Vogt and Christoph Leitner for their advice and help.

\bibliographystyle{IEEEtran}
\bibliography{bib}

\begin{thebibliography}{10}
\providecommand{\url}[1]{#1}
\csname url@samestyle\endcsname
\providecommand{\newblock}{\relax}
\providecommand{\bibinfo}[2]{#2}
\providecommand{\BIBentrySTDinterwordspacing}{\spaceskip=0pt\relax}
\providecommand{\BIBentryALTinterwordstretchfactor}{4}
\providecommand{\BIBentryALTinterwordspacing}{\spaceskip=\fontdimen2\font plus
\BIBentryALTinterwordstretchfactor\fontdimen3\font minus \fontdimen4\font\relax}
\providecommand{\BIBforeignlanguage}[2]{{%
\expandafter\ifx\csname l@#1\endcsname\relax
\typeout{** WARNING: IEEEtran.bst: No hyphenation pattern has been}%
\typeout{** loaded for the language `#1'. Using the pattern for}%
\typeout{** the default language instead.}%
\else
\language=\csname l@#1\endcsname
\fi
#2}}
\providecommand{\BIBdecl}{\relax}
\BIBdecl

\bibitem{statista}
\BIBentryALTinterwordspacing
Statista, ``Drone market size worldwide in selected years from 2021 to 2030,'' January 2024. [Online]. Available: \url{https://www.statista.com/statistics/1234521/worldwide-drone-market/}
\BIBentrySTDinterwordspacing

\bibitem{HASSANALIAN201799}
\BIBentryALTinterwordspacing
M.~Hassanalian and A.~Abdelkefi, ``Classifications, applications, and design challenges of drones: A review,'' \emph{Progress in Aerospace Sciences}, vol.~91, pp. 99--131, 2017. [Online]. Available: \url{https://www.sciencedirect.com/science/article/pii/S0376042116301348}
\BIBentrySTDinterwordspacing

\bibitem{zhang2024endtoend}
\BIBentryALTinterwordspacing
N.~Zhang, F.~Nex, G.~Vosselman, and N.~Kerle, ``End-to-end nano-drone obstacle avoidance for indoor exploration,'' \emph{Drones}, vol.~8, no.~2, 2024. [Online]. Available: \url{https://www.mdpi.com/2504-446X/8/2/33}
\BIBentrySTDinterwordspacing

\bibitem{lamberti2022tiny}
L.~Lamberti, V.~Niculescu, M.~Barciś, L.~Bellone, E.~Natalizio, L.~Benini, and D.~Palossi, ``Tiny-pulp-dronets: Squeezing neural networks for faster and lighter inference on multi-tasking autonomous nano-drones,'' in \emph{2022 IEEE 4th International Conference on Artificial Intelligence Circuits and Systems (AICAS)}, 2022, pp. 287--290.

\bibitem{mueller2023robust}
H.~Müller, V.~Niculescu, T.~Polonelli, M.~Magno, and L.~Benini, ``Robust and efficient depth-based obstacle avoidance for autonomous miniaturized uavs,'' \emph{IEEE Transactions on Robotics}, vol.~39, no.~6, pp. 4935--4951, 2023.

\bibitem{mcguire2019minimal}
K.~McGuire, C.~De~Wagter, K.~Tuyls, H.~Kappen, and G.~C. de~Croon, ``Minimal navigation solution for a swarm of tiny flying robots to explore an unknown environment,'' \emph{Science Robotics}, vol.~4, no.~35, 2019.

\bibitem{10423569}
K.~Kalenberg, H.~Müller, T.~Polonelli, A.~Schiaffino, V.~Niculescu, C.~Cioflan, M.~Magno, and L.~Benini, ``Stargate: Multimodal sensor fusion for autonomous navigation on miniaturized uavs,'' \emph{IEEE Internet of Things Journal}, pp. 1--1, 2024.

\bibitem{laurijssen2019flexible}
D.~Laurijssen, R.~Kerstens, G.~Schouten, W.~Daems, and J.~Steckel, ``A flexible low-cost biologically inspired sonar sensor platform for robotic applications,'' in \emph{2019 International Conference on Robotics and Automation (ICRA)}, 2019, pp. 9617--9623.

\bibitem{forouher2016sensor}
D.~Forouher, M.~G. Besselmann, and E.~Maehle, ``Sensor fusion of depth camera and ultrasound data for obstacle detection and robot navigation,'' in \emph{2016 14th International Conference on Control, Automation, Robotics and Vision (ICARCV)}, 2016, pp. 1--6.

\bibitem{zhmud2018application}
V.~Zhmud, N.~Kondratiev, K.~Kuznetsov, V.~Trubin, and L.~Dimitrov, ``Application of ultrasonic sensor for measuring distances in robotics,'' in \emph{Journal of Physics: Conference Series}, vol. 1015, no.~3.\hskip 1em plus 0.5em minus 0.4em\relax IOP Publishing, 2018, p. 032189.

\bibitem{rshen2019new}
M.~Shen, Y.~Wang, Y.~Jiang, H.~Ji, B.~Wang, and Z.~Huang, ``A new positioning method based on multiple ultrasonic sensors for autonomous mobile robot,'' \emph{Sensors}, vol.~20, no.~1, p.~17, 2019.

\bibitem{przybyla2023mass}
R.~J. Przybyla, S.~E. Shelton, C.~Lee, B.~E. Eovino, Q.~Chau, M.~H. Kline, O.~I. Izyumin, and D.~A. Horsley, ``Mass produced micromachined ultrasonic time-of-flight sensors operating in different frequency bands,'' in \emph{2023 IEEE 36th International Conference on Micro Electro Mechanical Systems (MEMS)}.\hskip 1em plus 0.5em minus 0.4em\relax IEEE, 2023, pp. 961--964.

\bibitem{palossi201964}
D.~Palossi, A.~Loquercio, F.~Conti, E.~Flamand, D.~Scaramuzza, and L.~Benini, ``A 64-mw dnn-based visual navigation engine for autonomous nano-drones,'' \emph{IEEE Internet of Things Journal}, vol.~6, no.~5, pp. 8357--8371, 2019.

\bibitem{duembgen2023blind}
F.~Dümbgen, A.~Hoffet, M.~Kolundžija, A.~Scholefield, and M.~Vetterli, ``Blind as a bat: Audible echolocation on small robots,'' \emph{IEEE Robotics and Automation Letters}, vol.~8, no.~3, pp. 1271--1278, 2023.

\end{thebibliography}

\end{document}